\documentclass[runningheads]{llncs}

\usepackage[T1]{fontenc}

\usepackage{graphicx}
\usepackage{color}
\usepackage{hyperref}

\usepackage{dcolumn}
\usepackage{multirow}
\usepackage{adjustbox}
\usepackage{enumitem}
\usepackage{amssymb}

\usepackage{listings}

\newcommand{\filledcirc}{\raisebox{0.5pt}{\scalebox{1.2}{$\bullet$}}}

\newcommand\availabilityurl{https://github.com/qzc438-research/ontology-hallucination}

\begin{document}

\title{OAEI-LLM-T: A TBox Benchmark Dataset for Understanding Large Language Model Hallucinations in Ontology Matching}
\titlerunning{OAEI-LLM-T Dataset}

\author{
Zhangcheng Qiang\inst{1}\orcidID{0000-0001-5977-6506} \and
Kerry Taylor\inst{1}\orcidID{0000-0003-2447-1088} \and
Weiqing Wang\inst{2}\orcidID{0000-0002-9578-819X} \and
Jing Jiang\inst{1}\orcidID{0000-0002-3035-0074}}

\authorrunning{Z. Qiang et al.}

\institute{Australian National University, Canberra, Australia \and Monash University, Melbourne, Australia}

\maketitle

\begin{abstract}
Hallucinations are often inevitable in downstream tasks using large language models (LLMs). To tackle the substantial challenge of addressing hallucinations for LLM-based ontology matching (OM) systems, we introduce a new benchmark dataset OAEI-LLM-T. The dataset evolves from seven TBox datasets in the Ontology Alignment Evaluation Initiative (OAEI), capturing hallucinations of ten different LLMs performing OM tasks. These OM-specific hallucinations are organised into two primary categories and six sub-categories. We showcase the usefulness of the dataset in constructing an LLM leaderboard for OM tasks and for fine-tuning LLMs used in OM tasks.
\end{abstract}

\section{Introduction}
\label{sec: introduction}

Hallucinations are innate features of large language models (LLMs), where LLMs tend to generate plausible sounding but factually incorrect content~\cite{zhang2023siren}. There are various causes of this phenomenon. Hallucinations can arise from data, where the data used for training may have misinformation, bias, or knowledge boundaries, or the sample alignment data used for fine-tuning may be of poor quality. Hallucinations are also caused by improper model training, including pre-training and fine-tuning using supervised fine-tuning (SFT) or reinforcement learning from human feedback (RLHF). Hallucinations may also occur due to imperfect decoding strategies, overconfidence, softmax bottlenecks, and reasoning failure during the inference process~\cite{huang2023survey}. While LLM hallucinations can be helpful for narrative writing~\cite{sui2024confabulation} and adversarial example generation~\cite{yao2023llm}, they have significant drawbacks and hamper the reliability of LLMs in many real-world scenarios~\cite{xu2024hallucination}.

Benchmarking LLM hallucinations is an essential but critical step to understanding the cause of LLM hallucinations and mitigating the effects. Most benchmark datasets focus on general-purpose question-answering (QA), leaving under-explored LLM hallucinations in domain-specific or task-specific scenarios. For example, hallucinations occur in ontology matching (OM) tasks. Unlike traditional general-purpose QA tasks, OM tasks aim to find aligned entities between two different ontologies, and this process requires a comprehensive understanding of the context based on the ontologies provided. For example, within the context of a research conference, the entity ``chair'' could be correctly mapped to the entity ``chairman'' when it refers to a specific meaning of the person who chairs the conference, but not when it refers to the meaning of furniture that could be used at the conference.

OM tasks are commonly executed by OM software systems. With the prevalence of LLMs, there is a trend of shifting classical expert systems~\cite{jimenez2011logmap1,jimenez2011logmap2,faria2013agreementmakerlight,faria2014agreementmakerlight,zhao2018matching,li2021combining} toward LLM-based OM systems~\cite{he2023exploring,norouzi2023conversational,hertling2023olala,qiang2023agent,giglou2024llms4om,amini2024towards}, in which LLMs not only provide strong background knowledge but also construct a language-instructed pipeline to link OM systems with open-world tools and APIs. LLM-based OM systems have advanced enormously in performing few-shot and complex OM tasks, but they have also inherited LLM hallucinations. In this paper, we introduce OAEI-LLM-T, a TBox benchmark dataset to understand LLM hallucinations in LLM-based OM systems. It evolves seven schema-matching datasets in the Ontology Alignment Evaluation Initiative (OAEI). We benchmark a total of ten different LLMs on OM tasks and classify the hallucinations that occur in these datasets. We showcase the use of OAEI-LLM-T to construct an LLM leaderboard for OM tasks, as well as using OAEI-LLM-T to fine-tune LLMs in order to improve performance on OM tasks.

The rest of the paper is organised as follows. Section~\ref{sec: background} introduces the OM task and hallucinations that occur in OM tasks. Section~\ref{sec: related work} reviews the related work in benchmarking LLM hallucinations. We describe the details of the dataset construction in Section~\ref{sec: dataset construction} and present use cases in Section~\ref{sec: use cases}. Section~\ref{sec: limitations} discusses the current limitations, and Section~\ref{sec: conclusion} concludes the paper.

\section{Background}
\label{sec: background}

\subsection{The Ontology Matching Task}

OM tasks aim to find aligned entities between two ontologies~\cite{euzenat2007ontology}. The ontology entities include classes, properties, and individuals. We call the matching that focuses on classes and properties TBox matching (a.k.a. schema matching), while the matching that focuses on individuals is ABox matching (a.k.a. instance matching). OM systems are developed to automate the matching process. The input of the OM system is a pair of ontologies, namely the source ontology ($O_s$) and the target ontology ($O_t$). Together with a pre-defined relation ($r$) and a similarity threshold ($s\in [0,1]$), the OM system produces a set of aligned entities where the level of confidence ($c$) is greater than the similarity threshold, formulated as alignment $A = \{(e1, e2, r, c) | e1 \in O_s, e2 \in O_t, s \leq c \leq 1\}$. If there is a gold standard reference (R), system performance can be measured by precision (Prec), recall (Rec), and F1 score (F1), defined as~\cite{do2003comparison}:

\begin{equation}
Prec = \frac{|A \cap R|}{|A|} \qquad Rec = \frac{|A \cap R|}{|R|} \qquad F_1 = \frac{2}{Prec^{-1} + Rec^{-1}}
\end{equation}

\subsection{LLM Hallucinations in Ontology Matching}

\begin{figure}[htbp]
\centering
\includegraphics[width=1\linewidth]{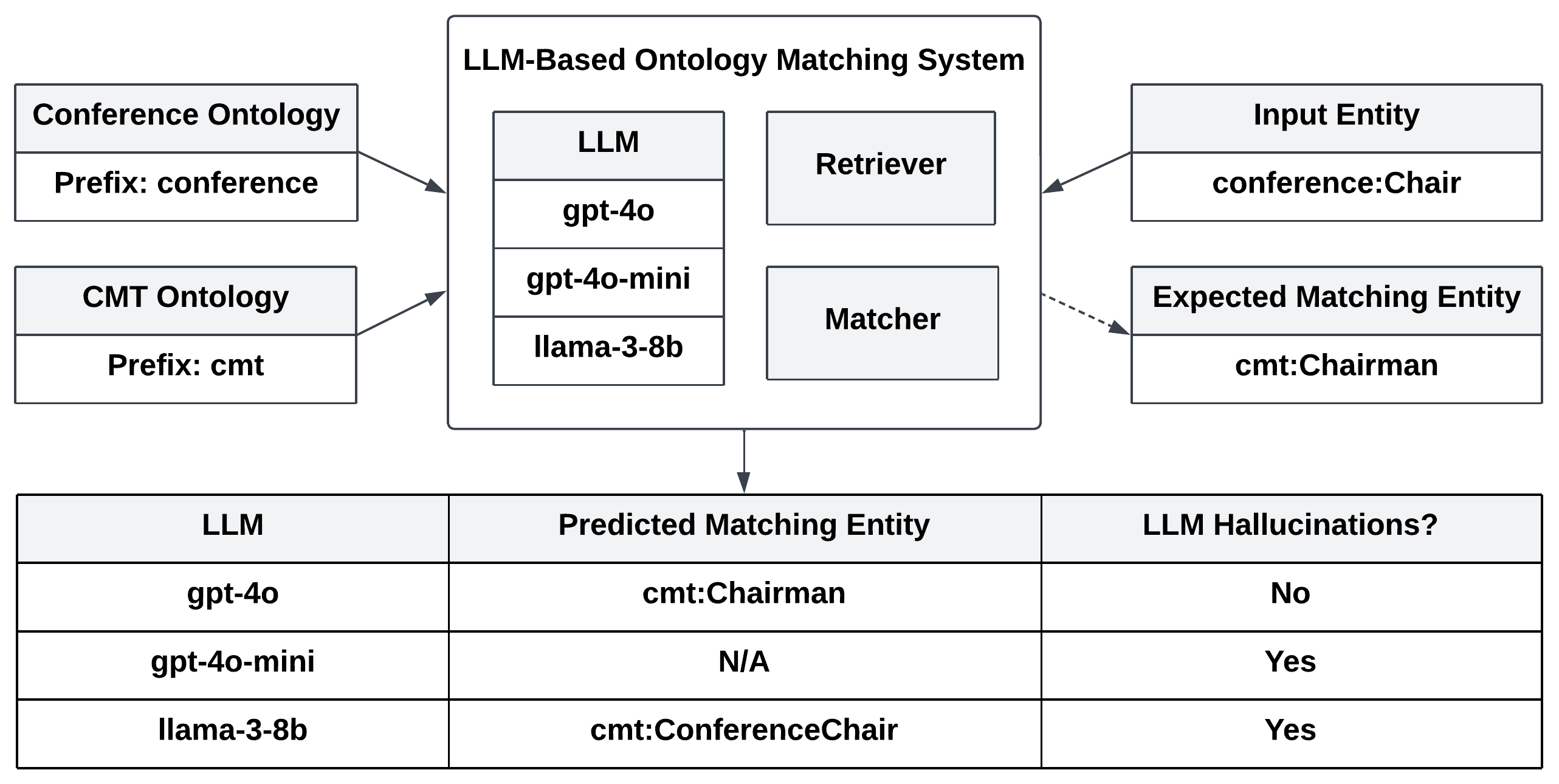}
\caption{An example of LLM hallucinations in LLM-based OM systems.}
\label{fig: om-hallucination}
\end{figure}

Fig.~\ref{fig: om-hallucination} illustrates an example of LLM hallucinations in LLM-based OM systems. Given a pair of ontologies (Conference Ontology and CMT Ontology), the LLM-based OM system is used to find the possible mappings between these two ontologies. For example, the input entity \emph{``conference:Chair''} is expected to match the entity \emph{``cmt:Chairman''}. An LLM-based OM system primarily consists of an LLM and two basic components utilising the LLM: the retriever and the matcher. Depending on the different LLMs chosen, the system may produce different mappings. For example, the predicted matching entity found by \emph{gpt-4o} is \emph{``cmt:Chairman''}. Since the predicted result is the same as the expected result as recorded in the reference, we determine that there is no LLM hallucination in this case. On the other hand, \emph{gpt-4o-mini} and \emph{llama-3-8b} are observed to have LLM hallucinations: the former cannot find any mappings, while the latter finds an incorrect mapping \emph{``cmt:ConferenceChair''}.

While an LLM-based OM system employs the different capabilities of LLMs, including information retrieval, tool use, and self-reflection, the cause of LLM hallucinations in LLM-based OM systems can be complex. Not only can the LLM be used in the system to cause hallucinations, but also the system components based on LLM (e.g. retriever and matcher) can cause hallucinations. It is possible that a missing or false mapping is caused by the retriever finding the wrong context information of the ontology, or due to the matcher misunderstanding the entity meaning or relations between entities. In general, an LLM-based OM system can either miss existing mappings in the reference or find incorrect mappings with respect to the reference. Both have a negative influence on system performance. Missing existing mappings will reduce recall, while finding incorrect mappings will reduce precision.

\section{Related Work}
\label{sec: related work}

A number of datasets have been developed to understand the LLM hallucinations that occur in various downstream tasks. Most of them focus on general-purpose question-answering (QA), including generative QA~\cite{cheng2023evaluating,li2023halueval,li2024dawn}, multi-choice QA~\cite{muhlgay2023generating}, and a combination of the above~\cite{lin2021truthfulqa,kasai2024realtime}. HaluEval-Wild~\cite{zhu2024halueval} evolves these datasets and develops real-world user queries. There are some domain-specific LLM hallucination benchmarks in medicine~\cite{pal2023med,zuo2024medhallbench}, biomedicine~\cite{seo2024dahl}, and manufacturing~\cite{sadat2023delucionqa}. There are also task-specific LLM hallucination benchmarks in dialogue summarisation~\cite{tang2024tofueval}, search engine augmentation~\cite{vu2023freshllms}, tool use~\cite{zhuang2023toolqa}, retrieval-augmented generation (RAG)~\cite{niu2023ragtruth}, and knowledge graphs (KGs)~\cite{yu2023kola}. BIG-bench~\cite{srivastava2023beyond} is a comprehensive dataset that contains more than 200 tasks. In addition to traditional natural language processing (NLP) tasks, it also contains benchmarks related to logic, mathematics, code, and advanced questions for understanding humans and the world. However, to the best of our knowledge, no datasets are currently available for LLM hallucinations that occur in OM tasks.

\section{Dataset Construction}
\label{sec: dataset construction}

\begin{figure}[htbp]
\centering
\includegraphics[width=\linewidth]{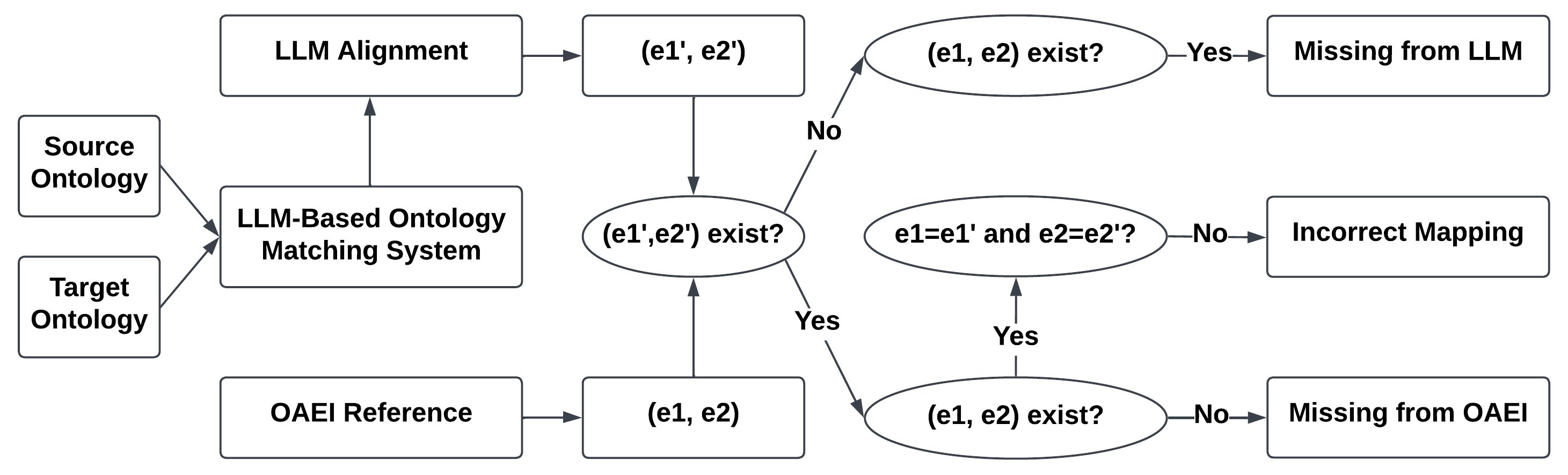}
\caption{An overview of the dataset construction.}
\label{fig: dataset-construction}
\end{figure}

Fig.~\ref{fig: dataset-construction} illustrates an overview of the dataset construction. Given a source ontology and a target ontology, the LLM-based OM system generates the alignment (i.e. a set of predicted mappings) using different LLMs. The LLM Alignment ($A_{llm}$) is compared with the OAEI Reference ($R_{oaei}$). We consider the cases that a mapping is \emph{Missing from LLM} or \emph{Missing from Reference} as follows. For some $(e1, e2) \in R_{oaei}$ where there is no $(e1',e2') \in A_{llm}$ such that either $e1=e1'$ or $e2 = e2'$ but not both $e1=e1'$ and $e2=e2'$, we consider the mapping $(e1, e2)$ to be \emph{Missing from LLM}. Correspondingly,  for some $(e1', e2') \in A_{llm}$ where there is no $(e1',e2') \in R_{oaei}$ such that either $e1=e1'$ or $e2 = e2'$ but not both $e1=e1'$ and $e2=e2'$, we consider the mapping $(e1', e2')$ to be \emph{Missing from Reference}. Otherwise, in the case that for some $(e1, e2) \in R_{oaei}$ there is an $(e1', e2') \in A_{llm}$  such that $e1 = e1'$  or $e2 = e2'$, but not both $e1=e1'$ and $e2=e2'$, we consider $(e1', e2')$ to be \emph{Incorrect Mapping}. There is no hallucination if $e1=e1'$ and $e2=e2'$ (i.e. exactly matched) or when both mappings are missing (i.e. exactly unmatched).

\subsection{Selected OAEI Reference}

The OAEI campaign contains a collection of OM datasets from different domains. In this study, we focus on understanding LLM hallucinations in the TBox. Therefore, the selected datasets are schema-matching datasets. We analyse and match classes and properties within each pair of ontologies but exclude their individuals.Table~\ref{tab: OAEI track} lists the selected OAEI tracks for the OAEI Reference.

\begin{table}[htbp]
\centering
\renewcommand\arraystretch{1.2}
\tabcolsep=0.15cm
\caption{Selected OAEI tracks for OAEI Reference.}
\label{tab: OAEI track}
\begin{tabular}{|l|l|l|c|}
\hline
\multirow{1}{*}{\textbf{Name}} & \multirow{1}{*}{\textbf{Domain}} & \multirow{1}{*}{\textbf{Main Challenge}} & \multirow{1}{*}{\textbf{Alignments}} \\ \hline
\multicolumn{1}{|l|}{anatomy}       & Human and Mouse Anatomy           & Domain Knowledge      & 1     \\ \hline
\multicolumn{1}{|l|}{conference}    & Research Conference               & Few-shot Prediction   & 24    \\ \hline
\multicolumn{1}{|l|}{food}          & Food Nutritional Composition      & Complex Logic         & 1     \\ \hline
\multicolumn{1}{|l|}{biodiv}        & Biodiversity and Ecology          & Domain Knowledge      & 9     \\ \hline
\multicolumn{1}{|l|}{mse}           & Materials Science \& Engineering  & Domain Knowledge      & 3     \\ \hline
\multicolumn{1}{|l|}{commonkg}      & Common Knowledge Graphs           & Polysemous Words      & 3     \\ \hline
\multicolumn{1}{|l|}{dh}            & Digital Humanities                & Domain Knowledge      & 8     \\ \hline
\end{tabular}
\end{table}

\subsection{Selected Ontology Matching System}

We select Agent-OM~\cite{qiang2023agent}, an agent-powered LLM-based OM system, to generate the system alignments for benchmarking. The system uses large pre-trained foundation LLMs without any fine-tuning as the back-end. This avoids the fine-tuning process to bring extra hallucinations into the matching results. Agent-OM currently supports several LLMs and thus allows the datasets to capture the behaviours of a diversity of LLMs. An agent-based system also brings automation with minimum human effort to handle large-scale OAEI datasets. For hyperparameter settings, we use the general optimal setting described in the original paper~\cite{qiang2023agent}, that is, similarity\_threshold = 0.90 and top@k = 3. For text embedding, we choose the OpenAI embedding model text-embedding-ada-002~\cite{openaiembeddingmodels}. The default length of the embedding vector is 1536.

\subsection{Selected Large Language Models}

Table~\ref{tab: llm} lists the selected LLMs within Agent-OM. We choose a total of ten models from six different families. The GPT and Claude models are API-accessed commercial LLMs, whereas the Llama, Qwen, Gemma, and ChatGLM models are open-source LLMs. Considering API cost and computational efficiency, we choose medium and small sizes of API-accessed commercial LLMs and open-source models ranging from 7-9 billion parameters. These models can be run on a single NVIDIA GeForce RTX 4090 graphics processing unit.

\begin{table}[htbp]
\centering
\renewcommand\arraystretch{1.2}
\tabcolsep=0.15cm
\caption{Selected large language models within Agent-OM.}
\label{tab: llm}
\begin{tabular}{|c|l|c|l|}
\hline
\multicolumn{1}{|c|}{\textbf{Family}} & \multicolumn{1}{|l|}{\textbf{Model}} & \multicolumn{1}{|c|}{\textbf{Size}}
& \multicolumn{1}{|l|}{\textbf{Version}}  \\ \hline
\multirow{2}{*}{GPT}    & $\blacktriangle$~gpt-4o           & N/A       & gpt-4o-2024-05-13                     \\ \cline{2-4}
                        & $\blacktriangle$~gpt-4o-mini      & N/A       & gpt-4o-mini-2024-07-18                \\ \cline{1-4}
\multirow{2}{*}{Claude} & $\blacktriangle$~claude-3-sonnet  & N/A       & claude-3-sonnet-20240229              \\ \cline{2-4}
                        & $\blacktriangle$~claude-3-haiku   & N/A       & claude-3-haiku-20240307               \\ \cline{1-4}
\multirow{2}{*}{Llama}  & \filledcirc~llama-3-8b            & 4.7 GB    & Ollama Model ID: 365c0bd3c000         \\ \cline{2-4}
                        & \filledcirc~llama-3.1-8b          & 4.9 GB    & Ollama Model ID: 46e0c10c039e         \\ \cline{1-4}
\multirow{2}{*}{Qwen}   & \filledcirc~qwen-2-7b             & 4.4 GB    & Ollama Model ID: dd314f039b9d         \\ \cline{2-4}
                        & \filledcirc~qwen-2.5-7b           & 4.7 GB    & Ollama Model ID: 845dbda0ea48         \\ \cline{1-4}       
\multirow{1}{*}{Gemma}  & \filledcirc~gemma-2-9b            & 5.4 GB    & Ollama Model ID: ff02c3702f32         \\ \cline{1-4}
\multirow{1}{*}{GLM}    & \filledcirc~glm-4-9b              & 5.5 GB    & Ollama Model ID: 5b699761eca5         \\ \hline
\multicolumn{4}{c}{Triangle ($\blacktriangle$): API-accessed models; Circle ($\bullet$): open-source models.}   \\
\end{tabular}
\end{table}

\subsection{Categories of LLM Hallucinations in Ontology Matching}

LLM hallucinations in general can be categorised into two primary types: factuality and faithfulness. Factuality hallucination refers to LLM output inconsistent with real-world facts, whereas faithfulness hallucination refers to LLM output inconsistent with instruction, context, or logic~\cite{huang2023survey}. However, the traditional classification of LLM hallucinations is not feasible for OM tasks. Unlike QA tasks, the matching result of OM tasks is a set of paired entities, where these mappings are domain-specific and based on the input ontologies and domain knowledge, and the justification of these mappings is based on the reference provided by domain experts. 

Table~\ref{tab: om-hallucination-category} illustrates the categories of LLM hallucinations in OM tasks with examples. We introduce two primary categories \emph{Missing} and \emph{Incorrect} with six sub-categories. The \emph{Missing} category means that a mapping is missing either from LLM or from the OAEI reference, while the \emph{Incorrect} category means that a related mapping exists in LLM and OAEI, but they do not agree. There are four sub-categories under incorrect mappings. \emph{Align-Up Mapping} means a source entity is mapped to an entity in the target that is a superclass of the correct entity in the target as determined by the reference, whereas \emph{Align-Down Mapping} means an entity is mapped to its subclass. Moreover, an incorrect mapping can be a \emph{False Mapping} that is factually wrong, but it is also possible to be a \emph{Disputed Mapping} that can be arguably grounded in a certain real-world fact. Observation, coupled with discussion with the OAEI community, leads us to recognise that so-called  ``gold standard'' references can be faulty.

For these mappings in the \emph{Incorrect} category, we use an LLM arbiter to determine whether the mapping is assigned to the four sub-categories. Prompt Template~\ref{lst: judge} shows the prompt template used by the LLM arbiter. The context refers to the specific domain of the track. For example, the context for the conference track is ``research conference''. The LLM arbiter uses the GPT model gpt-4o~\cite{gpt-4o} as the back-end, as it is the best-performing model with minimum LLM hallucinations~\cite{qiang2023agent}. The temperature is set to 0 to minimise the random output.

\begin{table}[htbp]
\centering
\renewcommand\arraystretch{1.2}
\tabcolsep=0.15cm
\caption{Categories of LLM hallucinations in ontology matching task.}
\label{tab: om-hallucination-category}
\begin{adjustbox}{width=1\columnwidth,center}
\begin{tabular}{|c|c|c|c|}
\hline
\multicolumn{1}{|c|}{\multirow{2}{*}{\textbf{Category}}} & \multicolumn{1}{c|}{\multirow{2}{*}{\textbf{Sub-Category}}} & \multicolumn{2}{c|}{\textbf{Example}} \\ \cline{3-4}
\multicolumn{1}{|c|}{} & \multicolumn{1}{c|}{} & \multicolumn{1}{c|}{\textbf{Expected Matching}} & \multicolumn{1}{c|}{\textbf{Predicted Matching}} \\ \hline
\multirow{2}{*}{Missing}            & Missing from LLM          & Chairman $\equiv$ Chair
                                                                & N/A                                   \\ \cline{2-4}
                                    & Missing from Reference    & N/A
                                                                & Chairman $\equiv$ Chair               \\ \hline
\multirow{4}{*}{Incorrect}          & Align-Up Mapping          & Chairman $\equiv$ Chair     
                                                                & ConferenceRole $\equiv$ Chair         \\ \cline{2-4}
                                    & Align-Down Mapping        & Chairman $\equiv$ Chair        
                                                                & SessionChair $\equiv$ Chair           \\ \cline{2-4}
                                    & False Mapping             & Chairman $\equiv$ Chair    
                                                                & ConferenceFurniture $\equiv$ Chair    \\ \cline{2-4}
                                    & Disputed Mapping          & Chairman $\equiv$ Chair      
                                                                & ConferenceChair $\equiv$ Chair        \\ \hline
\end{tabular}
\end{adjustbox}
\end{table}

\begin{lstlisting}[label=lst: judge, caption=Determine the sub-categories of incorrect mappings.]
LLM-generated label: {predict_entity}
Intended label: {true_entity}
Context: {context}
Choose an answer from 1-4 within the context. Give a short explanation.
1. False-mapping: LLM-generated label is irrelevant to intended label.
2. Disputed-mapping: LLM-generated label is relevant to intended label.
3. Align-up: LLM-generated label is superclass/property of intended label.
4. Align-down: LLM-generated label is subclass/property of intended label.
\end{lstlisting}

\section{Use Case Demonstration}
\label{sec: use cases}

\subsection{LLM Leaderboard for Ontology Matching}

The behaviours of various LLMs performing OM tasks vary. If we consider determining an uncertain mapping as a decision process, LLMs are shown to have some behaviours similar to those of humans in risk aversion. Some models tend to be risk-avoiding (i.e. withdrawing the uncertain mappings and therefore becoming missing from LLM), and others tend to be risk-seeking (i.e. exploring the uncertain mappings aggressively and therefore missing from Reference). Benchmarking these would be useful for selecting the appropriate models applied in different domains. For example, conservative LLMs tend to withdraw uncertain mappings in behaviour that may be appropriate for biomedical domains requiring high matching correctness, while aggressive LLMs tend to generate incorrect mappings demonstrating behaviour appropriate for general domains where post-hoc correction is expected to be done inexpensively by humans.

\begin{figure}[!t]
\centering
\includegraphics[width=0.243\textwidth]{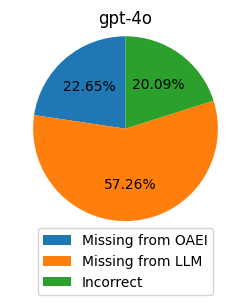}
\includegraphics[width=0.243\textwidth]{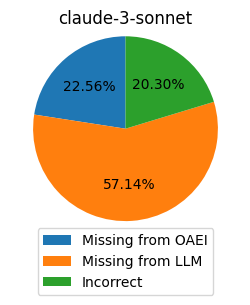}
\includegraphics[width=0.243\textwidth]{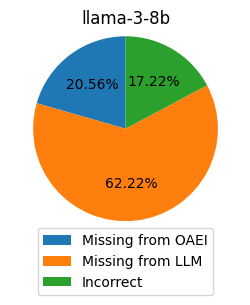}
\includegraphics[width=0.243\textwidth]{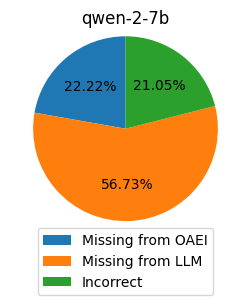}
\includegraphics[width=0.243\textwidth]{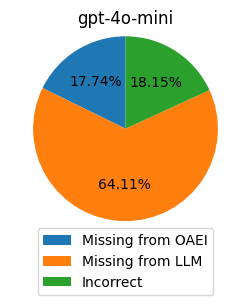}
\includegraphics[width=0.243\textwidth]{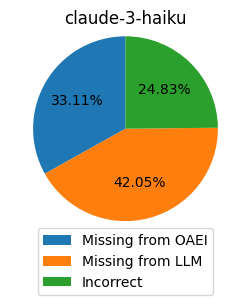}
\includegraphics[width=0.243\textwidth]{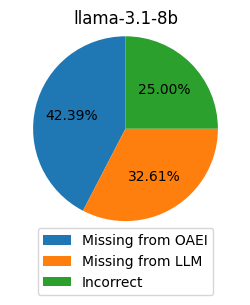}
\includegraphics[width=0.243\textwidth]{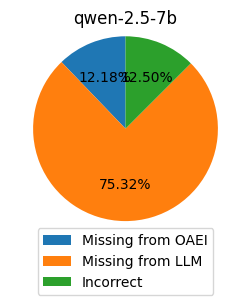}
\includegraphics[width=0.243\textwidth]{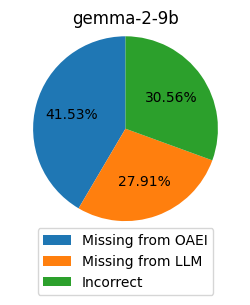}
\includegraphics[width=0.243\textwidth]{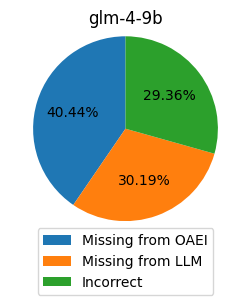}
\caption{Proportion of hallucination categories over a range of LLMs.}
\label{fig: compare-category}
\end{figure}

\begin{figure}[!t]
\centering
\includegraphics[width=0.243\textwidth]{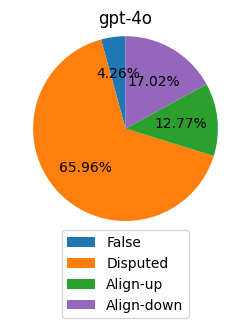}
\includegraphics[width=0.243\textwidth]{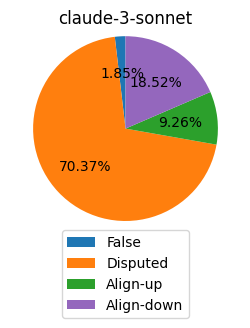}
\includegraphics[width=0.243\textwidth]{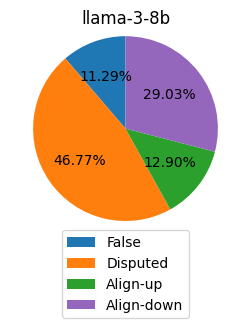}
\includegraphics[width=0.243\textwidth]{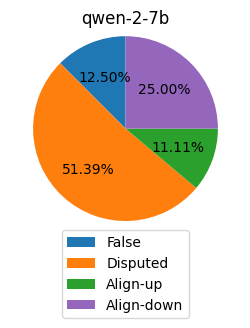}
\includegraphics[width=0.243\textwidth]{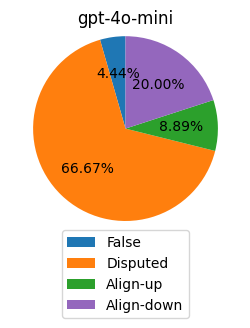}
\includegraphics[width=0.243\textwidth]{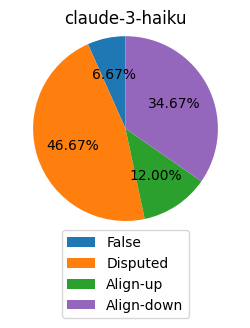}
\includegraphics[width=0.243\textwidth]{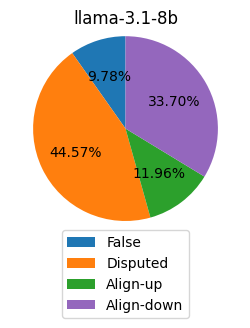}
\includegraphics[width=0.243\textwidth]{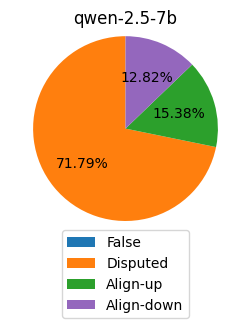}
\includegraphics[width=0.243\textwidth]{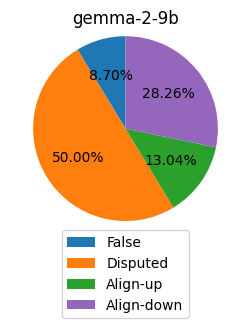}
\includegraphics[width=0.243\textwidth]{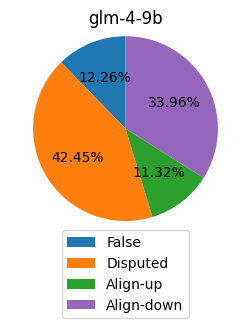}
\caption{Proportion of sub-categories of incorrect mappings over a range of LLMs.}
\label{fig: compare-incorrect-subcategory}
\end{figure}

Fig.~\ref{fig: compare-category} shows the proportion of hallucination categories in different LLMs in the OAEI Anatomy Track. For uncertain mappings, different LLMs tend to apply different strategies. We can see that the GPT, Claude, and Qwen models have more mappings missing from LLMs, while the Gemma and ChatGLM models have more mappings missing from Reference. Interestingly, Llama models have shown different behaviours across different versions. While llama-3-8b tends to have more missing from LLMs, llama-3.1-8b tends to have more missing mappings from OAEI. Fig.~\ref{fig: compare-incorrect-subcategory} shows the proportion of sub-categories of incorrect mappings in different LLMs in the OAEI Anatomy Track. We further evaluate the relevance of these incorrect mappings. LLMs tend to generate relevant mappings, and there are very few false mappings. These relevant mappings are commonly disputed and align-down mappings, with a few align-up mappings.

\subsection{LLM Fine-Tuning for Ontology Matching}

Fine-tuning is an effective way to customise LLMs for a domain-specific task. There are two dominant strategies used in fine-tuning for LLMs: supervised fine-tuning (e.g. SFT) and preference fine-tuning (e.g. DPO~\cite{rafailov2023direct} or ORPO~\cite{hong2024orpo}). The former is useful for injecting domain knowledge into LLMs, and the latter is helpful to customise the style and tendencies of LLMs for a specific task. 

The OAEI-LLM-T dataset can be used to generate the fine-tuning dataset. The key concept is to teach LLMs to generate answers according to the OAEI Reference and avoid generating hallucinated answers captured in the OAEI-LLM-T dataset. While state-of-the-art fine-tuning libraries (e.g. TRL~\cite{vonwerra2022trl}) are commonly designed for QA tasks, the main challenge here is to transform an OM task to comply with the QA format.

Prompt Template~\ref{lst: question} shows the prompt template to generate the question and its contrasting answers. This task is conducted by an LLM debater that uses the advanced GPT model gpt-4o as the back-end, with the temperature set to 0. The fundamental component of the template is a binary question: ``Is entity e1 equivalent to entity e2?'' It is essential to provide detailed information about each entity, including the syntactic, lexical, and semantic information for each entity. We reuse the corresponding LLM-based tools in Agent-OM to retrieve these pieces of information from the ontologies. For the supervised fine-tuning dataset, we ask the LLM debater to provide only a chosen answer. For the preference fine-tuning dataset, on the other hand, we ask the LLM debater to provide an additional rejected answer. Assuming that the chosen answer gives a ``yes'' statement and explanation to the question, then the rejected answer gives a ``no'' statement and explanation to the question.

\begin{lstlisting}[label=lst: question, caption=Generate the question and its contrasting answers.]
Is entity e1 equivalent to entity e2? Consider the following:
Syntactic information for e1: syntactic_retriever(e1)
Lexical information for e1: lexical_retriever(e1)
Semantic information for e1: semantic_retriver(e1)
Syntactic information for e2: syntactic_retriever(e2)
Lexical information for e2: lexical_retriever(e2)
Semantic information for e2: semantic_retriver(e2)
Generate a positive answer to state "Yes, e1 is equivalent to e2".
Generate a negative answer to state "No, e1 is not equivalent to e2".
\end{lstlisting}

Table~\ref{tab: answer-fine-tuning} shows the generated answers in the fine-tuning dataset. We choose a positive answer for \emph{Missing from LLM} and choose a negative answer for \emph{Missing from OAEI}. For \emph{Incorrect Mappings}, the prompt will be iterated twice to find two statements: one for the true mappings captured by the Reference, and one for the false mappings captured by the LLM. Note that incorrect mappings used for fine-tuning LLMs include the sub-categories of align-up, align-down, and false mappings. Disputed mappings require an additional check to determine the ground truth for positive and negative answers.

\begin{table}[htbp]
\centering
\renewcommand\arraystretch{1.2}
\tabcolsep=0.15cm
\caption{Generated answers in the fine-tuning dataset.}
\label{tab: answer-fine-tuning}
\begin{adjustbox}{width=1\columnwidth,center}
\begin{tabular}{|c|c|c|c|}
\hline
\multicolumn{1}{|c|}{\textbf{Category}} & \multicolumn{1}{|c|}{\textbf{Sub-Category}} & \multicolumn{1}{c|}{\textbf{Chosen Answer}}   & \multicolumn{1}{c|}{\textbf{Rejected Answer}} \\ \hline
\multirow{2}{*}{Missing} & Missing from LLM & Positive                                      & Negative                                      \\  \cline{2-4}
& Missing from Reference                      & Negative                                      & Positive                                      \\ \hline
\multirow{2}{*}{Incorrect} & Align-Up \& Align-Down  & Positive (from Reference)                     & Negative (from Reference)                     \\ \cline{3-4}
& \& False Mappings                           & Negative (from LLM)                      & Positive (from LLM)                      \\ \hline
\end{tabular}
\end{adjustbox}
\end{table}

While splitting a dataset into training and testing is a common practice in machine-learning-based systems, the OAEI datasets do not have this setting and the fine-tuning datasets are retrieved from the complete dataset. We advise users of our fine-tuning dataset to exclude mappings obtained from the respective reference alignments when planning to fine-tune for OAEI alignment challenges.

\section{Limitations}
\label{sec: limitations}

In this study, the hallucinations we captured are the significant ones generated by the LLM-based OM system. We cannot capture the complete set of potential LLM hallucinations (and it is not necessary) because there are too many and most of them are unhelpful. We observe that for each entity in the source ontology there are potentially at most $(n-1)$ mappings, where $n$ is the number of entities in the target ontology. Many of these are consistently non-matched across multiple LLMs. Therefore, there is no need to capture these false mappings in the dataset because they are not practically helpful for training and fine-tuning.

We use LLMs to classify different types of LLM hallucinations and curate fine-tuning datasets for OM tasks. Although we use the most advanced models (e.g. gpt-4o) for these tasks, we cannot guarantee that there is no bias in the LLM arbiter and debater. For the sake of fair comparisons, we use a standardised hyperparameter setting of Agent-OM to curate the input data for the benchmark across different tracks and LLMs. This setting is based on our experience and may not be optimal for certain tracks and LLMs.

\section{Conclusion}
\label{sec: conclusion}

In this paper, we introduce a dataset OAEI-LLM-T. We extend the original proposal OAEI-LLM~\cite{qiang2024oaei}, as presented at the ISWC 2024 Special Session on Harmonising Generative AI and Semantic Web Technologies (HGAIS 2024), with comprehensive implementation details on the dataset construction and showcase the use of the dataset in two real-world applications: (1) Use OAEI-LLM-T to create an LLM leaderboard to support LLM selection for OM tasks. (2) Convert OAEI-LLM-T to a fine-tuning dataset to optimise the LLMs used in OM tasks.

OM is a long-term problem studied for the Semantic Web. With the popularity of LLMs, traditional expert systems are currently moving to modern LLM-based OM systems. OAEI-LLM aims to capture LLM hallucinations that occur in various OM tasks to better understand the OM systems built on LLMs. We hope this contributes to future progress in OM where we envisage in the near future that LLM-informed OM will exceed the performance of human matchers in quality, not only in cost and speed.

\paragraph*{Resource Availability Statement:} 

OAEI-LLM-T is the first dataset released under the proposal of the OAEI-LLM series~\cite{qiang2024oaei}, focusing on LLM hallucinations in the TBox (class and property) matching. The dataset and the source code have been made available at \url{\availabilityurl}. The baseline ontology matching system Agent-OM can be downloaded from \url{https://github.com/qzc438/ontology-llm}. The OAEI datasets can be downloaded from the OAEI Matching EvaLuation Toolkit (MELT) at \url{https://dwslab.github.io/melt/track-repository} (retrieved January 1, 2025). The OAEI data policy can be found in \url{https://oaei.ontologymatching.org/doc/oaei-deontology.2.html}.

\begin{credits}
\subsubsection{\ackname} The authors thank the organisers of the ISWC 2024 Special Session on Harmonising Generative AI and Semantic Web Technologies (HGAIS 2024) for organising the workshop and group discussion. The authors also thank the participants of the group discussion (Daniel Garijo, Ernesto Jimenez-Ruiz, Romuald Esdras Wandji, and Valentina Tamma) for providing insightful comments. The authors also thank the Commonwealth Scientific and Industrial Research Organisation (CSIRO) for supporting this project.
\end{credits}

\bibliographystyle{splncs04}
\bibliography{mybibliography}

\end{document}